\documentclass[10pt,twocolumn,twoside]{IEEEtran} 
\usepackage{cite}

\ifCLASSINFOpdf
  \usepackage[pdftex]{graphicx}
  \graphicspath{{./figures/}}
  \DeclareGraphicsExtensions{.pdf,.jpeg,.png}
\else
  \usepackage[dvips]{graphicx}
  \graphicspath{{./figures/}}
  \DeclareGraphicsExtensions{.eps, .jpeg,.png}
\fi

\usepackage{epstopdf}
\usepackage{epsfig}
\usepackage{color}
\usepackage{overpic}

\ifCLASSOPTIONcompsoc
  \usepackage[caption=false,font=normalsize,labelfont=sf,textfont=sf]{subfig}
\else
  \usepackage[caption=false,font=footnotesize]{subfig}
\fi

\usepackage[cmex10]{amsmath}
\usepackage{bbm}
\usepackage{amssymb}
\usepackage{times}
\usepackage{multirow}
\usepackage{float}
\usepackage{xspace}

\newcommand*{\myfont}{\fontfamily{pzc}\selectfont}
\makeatletter
\DeclareRobustCommand\onedot{\futurelet\@let@token\@onedot}
\def\@onedot{\ifx\@let@token.\else.\null\fi\xspace}
\def\eg{e.g\onedot}

\def\etal{\emph{et al}\onedot}
\makeatother

\usepackage{algorithmic}

\usepackage{array}

\begin{document}

\title{Stereo on a Budget}

\author{Dana~Menaker,~Shai~Avidan}

\maketitle

\begin{abstract}
We propose an algorithm for recovering depth using less than two images. Instead of having both cameras send their entire image to the host computer, the left camera sends its image to the host while the right camera sends only a fraction $\epsilon$ of its image. The key aspect is that the cameras send the information without communicating at all. Hence, the required communication bandwidth is significantly reduced.

While standard image compression techniques can reduce the communication bandwidth, this requires additional computational resources on the part of the encoder (camera). We aim at designing a light weight encoder that only touches a fraction of the pixels. The burden of decoding is placed on the decoder (host).

We show that it is enough for the encoder to transmit a sparse set of pixels. Using only $1+\epsilon$ images, with $\epsilon$ as little as 2\% of the image, the decoder can compute a depth map. The depth map's accuracy is comparable to traditional stereo matching algorithms that require both images as input. Using the depth map and the left image, the right image can be synthesized. No computations are required at the encoder, and the decoder's runtime is linear in the images' size.
\end{abstract}


\IEEEpeerreviewmaketitle

\section{Introduction}

%

Stereo matching algorithms assume that both images are available for processing. This puts a burden on the host computer that must capture both images even though they are highly correlated with each other. Once captured, the host can recover the depth map of the scene and there are numerous algorithms for doing so.

Our goal is to minimize the communication cost between the cameras and the host and still be able to produce a depth map of the scene, as well as both images captured by the cameras. Our intent is to let the left camera transmit its image to the host and let the right camera transmit only a fraction $\epsilon$ of its image. The host uses the $1+\epsilon$ images to compute the depth map. Using the left image and the depth map, a high quality estimate of the right image can be generated. The most important aspect of our work is that the right camera cannot communicate with the left camera. What information should the right camera send to the host?

The right camera can use a standard image compression algorithm to reduce the communication bandwidth to the host but this, in turn, places a higher computational burden on the camera. Higher computational cost translates to higher battery consumption and we would like to avoid that as much as possible.

The scenario we envision is a group of people taking pictures of the same scene with multiple smartphones and uploading them to the cloud where the host can then run a stereo matching algorithm. Because all smartphones capture the same scene the images they capture are highly correlated. It is therefore a waste to let each smartphone compress and transmit highly correlated images.

As a first step toward reaching this goal we consider a simple stereo pair with two calibrated and synchronized cameras. The left camera transmits its image $I_1$ to the host and the right camera transmits an encoded image $\bar{I_2}$. Suppose that $\bar{I_2}$ is a low-resolution version of the original image $I_2$. Then the host must solve a super resolution problem where given a pair of images $(I_1,\bar{I_2})$ it must recover both the depth and an approximation ${\hat I}_2$ of the true high resolution image $I_2$.

This straightforward approach still requires the right camera to touch every pixel in $I_2$ in order to construct the low resolution $\bar{I_2}$. We argue that this is the worst possible choice. To understand why, take this approach to the extreme. Suppose the right camera can send only one pixel to the host and the value of this pixel would be the mean intensity of $I_2$. But because $I_1$ and $I_2$ are images of the same scene they are highly correlated and therefore their mean intensities are highly correlated. Using the mean intensity of $I_1$ would be a good enough approximation. We haven't gained much from sending the mean intensity of $I_2$. We give a better alternative.

Instead of sending a low resolution image of $I_2$ we sample a sparse grid (without smoothing) of $I_2$ and send it. The sparse grid keeps the high frequencies of $I_2$ at the cost of introducing aliasing and we use $I_1$ to resolve this problem. Our key insight is that even a small fraction of $I_2$ is sufficient to compute a disparity map by using Joint Bilateral Filter with $I_1$ serving as the guidance image. Once we have the depth map we can recover a high quality approximation ${\hat I}_2$ of $I_2$.

\section{Background}

There is inherent redundancy in a stereo image pair, and stereoscopic compression algorithms use this redundancy in order to encode the stereo pair efficiently, \eg \cite{StereoCodingProjection}. Most stereo compression techniques use disparity compensation, with one image serving as a reference and the other predicted using the reference image and the disparity field. The residual image can also be encoded for improved performance. However, these techniques require the knowledge of both stereo images at the encoder, unlike the scenario we address.

In our scenario we wish to encode a single image, without information about its stereo counterpart (expect that it exists). Furthermore, we would like the encoding to be as light as possible, and a sampling of the image seems attractive. The topic of image sampling has been studied extensively, and one particular sampling method, called Farthest Point Strategy (FPS) \cite{FarthestPointSampling}, aims at reducing the communication bandwidth, as we do. This method preserves the sampling uniformity, while being random and without adding the extra cost of transmitting each pixel's coordinates.

The redundancy in a stereo image pair is also utilized in 3D-TV applications, where different views of a real world scene can be synthesized from a monoscopic view and the associated per-pixel depth information \cite{Depth-image-based-rendering}.

Our work is also related to super resolution from multiple cameras where the goal is to recover a high resolution video from a collection of low resolution videos and high resolution still images \cite{SuperResolutionECCV2002}. The key difference is that in our case we choose what information to send and can therefore avoid sending redundant information.

Disparity estimation algorithms can be divided into {\em global} methods that solve a global optimization problem or {\em local} methods that estimate disparity values for each pixel independently. An extensive survey of methods can be found in \cite{Szeliski-Scharstein}.

Local methods compute, for every pixel in the reference image, the cost for a range of disparity values. The disparity value with the lowest cost is assigned to that pixel. Because a single pixel may not be robust to noise, it's common to aggregate information in a neighborhood. One way to do that is to use a bilateral filter (BF) \cite{Yoon-Kweon-2006}. Instead of aggregating information over a rectangular window, BF is used to respect edge boundaries in the aggregation step.

The bilateral filter was developed as an edge preserving filter, where the weight of pixels is based on space-range distance. See \cite{BilateralFilter09} for a review of the topic.
An interesting extension of BF is the realization that the weights of the filter need not come from the input image itself but rather from some guidance image. For example, the case of Flash/No Flash photography. The No Flash image, that has warm colors but a lot of noise, is filtered with the guidance of the Flash image that has cold colors, but is less noisy \cite{Eisemann:2004:FPE,Petschnigg:2004:DPF}. The same principle was applied in Joint Bilateral Upsampling \cite{joint-bilateral-upsampling} where a high resolution image served as a guide when upsampling a low resolution image. This led to the general idea of Guided Image Filter \cite{He:2010:GIF} that also offers an exact algorithm that is linear in the size of the image.

Edge preserving filtering has also been used in the context of depth maps to estimate a high resolution depth map from a low resolution active 3D time-of-flight camera and a high resolution RGB image \cite{HighQualityDepthMapUpsampling}. This setup, however, is quite different from ours. It is an active method while ours is passive. That is, the pixels we send do not carry accurate depth information as is the case of a ToF camera.

\section{Depth Estimation} \label{method}

In this section we first describe how to estimate the disparity map using $1+\epsilon$ images.

Consider a stereo pair where the left camera sends its image $I_1$ to the host and the right camera sends an encoded image, $\bar{I}_2$, to the host. The host must recover the disparity map $D$ from $I_1$ and $\bar{I}_2$. The question is, what is the best $\bar{I}_2$ to use? We evaluate two encoding options.

The first, denoted {\myfont Downsample}, is to take $\bar{I}_2$ to be a downsampled version of $I_2$, which is equivalent to having an asymmetrical pair of cameras: one with a higher resolution than the other. This is appealing as it can reduce hardware costs. However, fine details are lost in this process, as we will later show.

The second, denoted {\myfont Sparse}, is to take $\bar{I}_2$ to contain sparse samples from $I_2$. In this scenario the camera capturing $I_2$ can have a high resolution, and the amount of information transmitted can vary according to the available bandwidth.

Each of these encoding schemes has it own upsides and downsides. Finally, we consider a {\myfont Hybrid} method, in which $\bar{I}_2$ contains sparse samples from $I_2$ as in {\myfont Sparse}, but the stereo matching is inspired by both approaches.

\subsection{Preliminaries}\label{Preliminaries}

Depth estimation algorithms often use the Disparity Space Image (DSI), which is calculated from two stereo images. The DSI is a 3D volume that assigns a cost to each pixel and disparity value, and the goal of Depth Estimation algorithms is to take the DSI as input and return a depth map as output. This involves choosing one, and only one, disparity value per pixel.

The DSI can be easily computed given a pair of stereo images $I_1$ and $I_2$. A commonly used cost measure is the sum of absolute differences. Formally, given images $I_1$ and $I_2$ we define the Disparity Space Image ${\cal D}$ to be:
\begin{equation}
{\cal D}(x,y,d) = |I_1(x,y) - I_2(x+d,y)|
\label{eq:dsi}
\end{equation}

Many stereo methods perform a cost aggregation step on the DSI. For example, Yang \cite{cvpr-12-qingxiong-yang} aggregates costs adaptively, based on pixel similarity, which is derived from $I_1$ in order to preserve edges. Hence, $I_1$ is an input to the aggregation step as well. Yang proposed a left-right consistency check that improves results, but requires $I_2$, and not just sparse samples of it.

\subsection{\myfont {\large Downsample}}\label{Downsample}

\begin{figure}[t]
\begin{center}
\includegraphics[width=\linewidth]{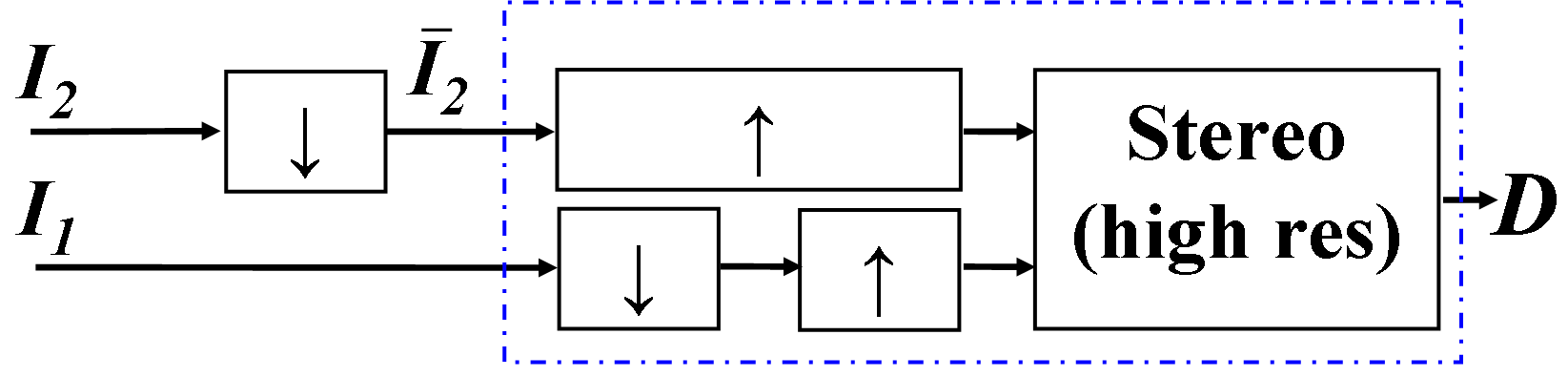} \\
(a) {\myfont Downsample} \\
\vspace{0.5em}
\includegraphics[width=\linewidth]{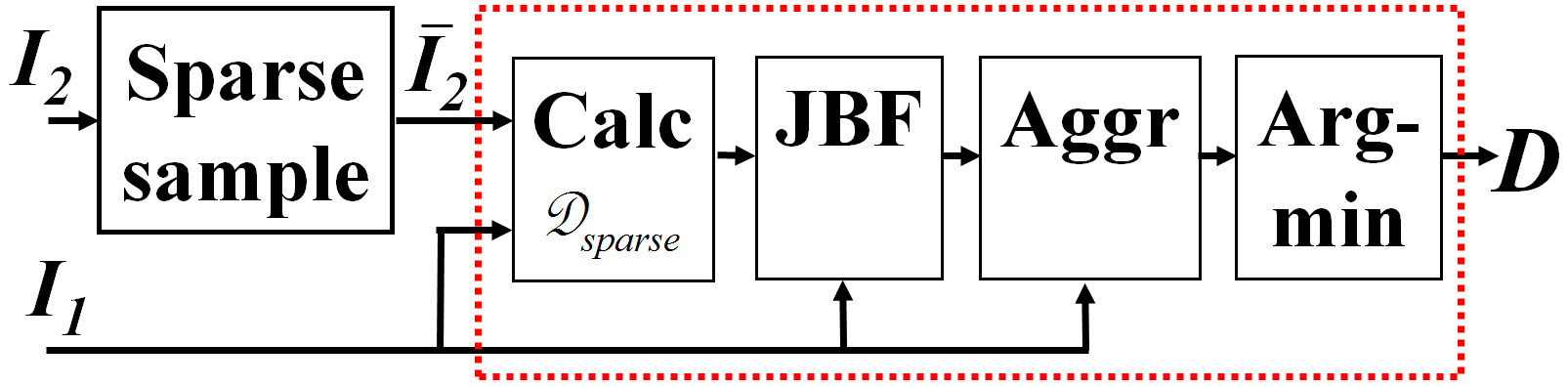} \\
(b) {\myfont Sparse} \\
\vspace{0.5em}
\includegraphics[width=\linewidth]{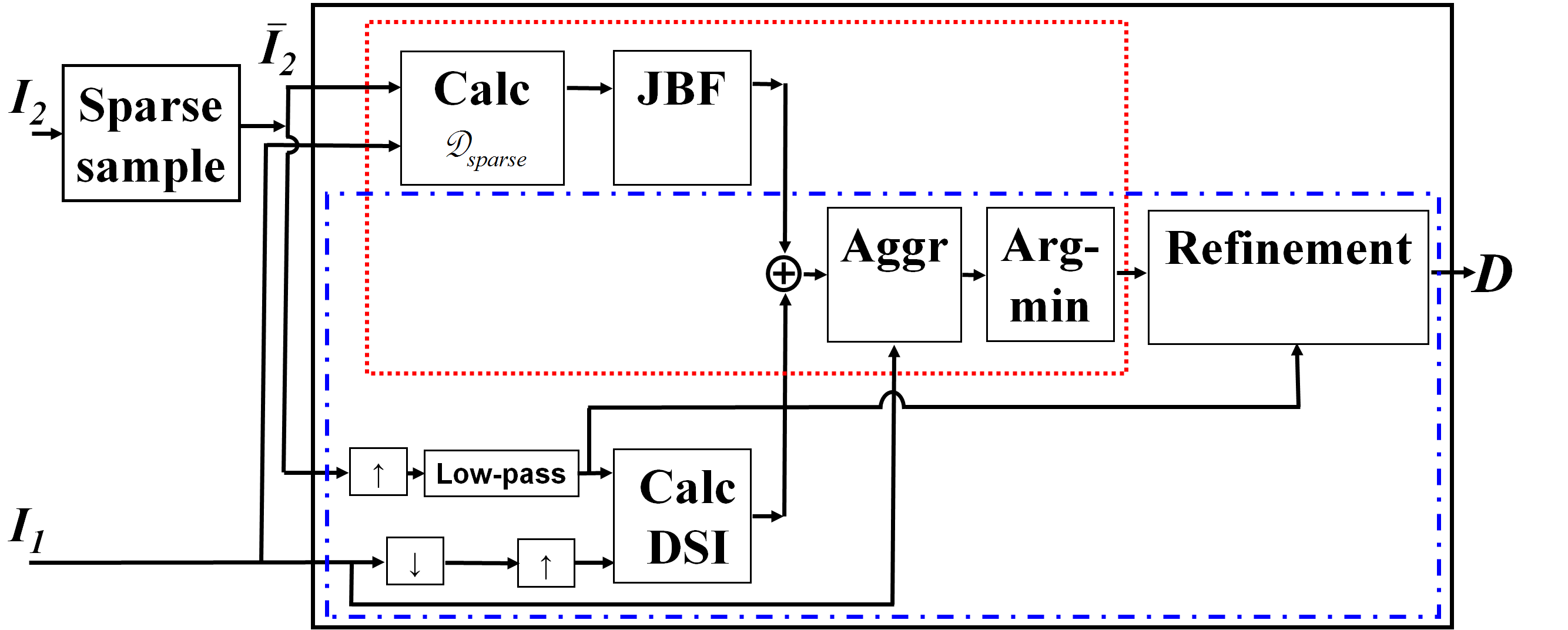} \\
(c) {\myfont Hybrid} \\
\vspace{0.5em}
\caption{Algorithm flowcharts. In (a) $\bar{I}_2$ denotes a downsampled version of $I_2$, in (b-c) it denotes a sparse sampling of $I_2$.}
\label{fig:Flowcharts}
\end{center}
\end{figure}

Let $\bar{I}_2$ denote a downsampled version of $I_2$. The {\myfont Downsample} algorithm is presented in figure \ref{fig:Flowcharts}(a). We upsample $\bar{I}_2$ to the original size and use it as input to a standard stereo method, which yields a high resolution disparity map. When doing so, it is crucial to smooth $I_1$ prior to the stereo matching, in order to match the spectral frequencies between the images. Otherwise the high frequencies in $I_1$, which are not present at the upsampled $\bar{I}_2$, will introduce noise to the matching.

\begin{figure}[tb]
\begin{center}
\begin{tabular}{cc}
\includegraphics[width=0.45\linewidth]{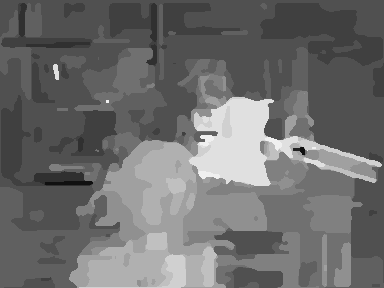} &
\includegraphics[width=0.45\linewidth]{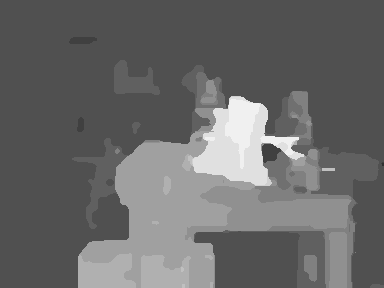} \\
(a) & (b) \\
\end{tabular}
\end{center}
\caption{Stereo matching results of tsukuba, where $\bar{I}_2$ is a downsampled version of $I_2$, with a scale factor 5 in each dimension. In (a) $\bar{I}_2$ is upsampled and the stereo matching is performed with the unprocessed $I_1$. In (b) $I_1$ is smoothed prior to the stereo matching. The stereo matching was done with \cite{cvpr-12-qingxiong-yang}. In (a), 14\% of the pixels in the disparity map have an error larger than 1, while in (b) only 8.18\% pixels have such an error.}
\label{fig:DisparityMapsDownsample}
\end{figure}

The importance of smoothing $I_1$ is demonstrated in figure \ref{fig:DisparityMapsDownsample}, where (a) is the matching result when $\bar{I}_2$ is a downsized version of $I_2$ by $5$ in each dimension, and (b) is the matching result with both images smoothed prior to matching (as shown in Fig. \ref{fig:Flowcharts}(a)). We used the stereo matching algorithm of \cite{cvpr-12-qingxiong-yang} in both cases. The result without smoothing $I_1$ has many discontinuities, caused by edges in the image which are not necessarily depth discontinuities, such as the shadow cast by the head on the table behind it, or the various folders and boxes on the shelves in the background. Those edges are not sharp in the upsampled $\bar{I}_2$, and therefore there is an ambiguity in the matching costs to the correct disparity value. The aggregation assumes correlation between depth and color, which is not a valid assumption in this case. Even if a different stereo matching algorithm, which includes a global optimization step (such as graph cut) was used, the weight on the smoothness term would have to be increased significantly, to the point of losing details in other areas.

The input to the stereo matcher is symmetric in the sense that both images have the same resolution. That enables us to perform the left-right consistency check suggested in \cite{cvpr-12-qingxiong-yang}, which improves the disparity map's accuracy. Since the images contain only low frequencies, we expect the disparity map to contain only low frequencies as well.

\subsection{\myfont {\large Sparse}}\label{Sparse}

Let $\bar{I}_2$ denote a sparse sampling of $I_2$. In {\myfont Sparse} the high frequencies in $I_2$ can be transmitted, contrary to {\myfont Downsample}. We wish to sample the image in a manner that will allow us to extract depth information in conjunction with $I_1$.

A progressive image sampling, which aims at minimizing the communication bandwidth, is described in \cite{FarthestPointSampling} (called Farthest Point Strategy - FPS). This sampling is random, however only the first pixel's coordinates should be transmitted, and the rest of the pixels' coordinates are determined by the previously transmitted pixels' intensities. This sampling can guarantee a uniform density, and it can also be adaptive to the content of the image, with a higher sample density in areas with finer details (denoted adaptive-FPS). Another advantage is that the irregular sampling corresponds to convolution of the signal with a wideband noise which reduces aliasing.

However, this sampling strategy requires a significant amount of computations, which contradicts our aim at designing a light weight encoder. Therefore, we sample $I_2$ on a uniform grid, requiring very little power and no computations at the camera.

The entire {\myfont Sparse} scheme is depicted in figure \ref{fig:Flowcharts}(b). First, ee use $\bar{I}_2$ and $I_1$ to calculate a sparse DSI (denoted ${\cal D}_{sparse}$) according to the following equation:
\begin{equation}
\resizebox{1\hsize}{!}{$ {\cal D}_{sparse}(x,y,d) = {\mathbbm{1}[ \bar{I_2}(x+d,y) ]} \cdot max \{ |I_1 (x,y) - \bar{I_2}(x+d,y)| , \delta \} $}
\end{equation}
where $\mathbbm{1}[ \cdot ]$ is an indicator function. Note that ${\cal D}_{sparse}$ is sparse and has many zero entries due to lack of data. In order to distinguish the case of a zero entry due to equal intensities from the case of missing data, we use $\delta = 10^{{-}6}$.

We then upgrade each layer of ${\cal D}_{sparse}(x,y,d)$ (for every $d$) using Joint Bilateral Filter, with $I_1$ as the guidance image. We denote the result of the filtering as $\hat{{\cal D}}$. In this process we exploit the correlation between color and disparity. Because $D$ and $I_1$ are expressed in the same coordinate system we do not have to estimate an intermediate motion field between them.

For the Bilateral Filter we use the fast implementation of \cite{Chaudhury:2011:FOB}, and we take into account the fact that the filtered DSI is sparse according to the following equation:
\begin{equation}
\resizebox{.9\hsize}{!}{$ \hat{{\cal D}}(x,y,d)= \tfrac {\displaystyle \sum_{(x',y') \in \Omega} { {\cal D}_{sparse}(x',y',d) F(x,y,x',y',I_1) } } { \displaystyle \sum_{(x',y') \in \Omega} { {\mathbbm{1}[ \cal D}_{sparse}(x',y',d) ] F(x,y,x',y',I_1)}} $}
\label{eq:sparse-dsi-filtering}
\end{equation}
where the filter $F(\cdot)$ is defined by
\begin{equation}
\resizebox{.9\hsize}{!}{$ F(\cdot)=f(\lVert (x,y){-}(x',y')\rVert)g( \lVert I_1(x,y){-}I_1(x',y') \rVert ) $}
\label{eq:sparse-dsi-filter-coeffs}
\end{equation}
where $f(\cdot)$ and $g(\cdot)$ are range kernels.

Given the estimated DSI $\hat{{\cal D}}$, we can use it instead of the full DSI in a stereo matching algorithm. As shown in Fig. \ref{fig:Flowcharts}(b), the aggregation and final disparity selection are done according to \cite{cvpr-12-qingxiong-yang}. Since $I_2$ is not available, we must skip the left-right consistency check.

\subsection{\myfont {\large Hybrid}}\label{Hybrid}

In {\myfont Downsample} the lower frequencies are transmitted, while in {\myfont Sparse} the high frequencies help preserve the details. In order to enjoy the best of both approaches, we wish to combine {\myfont Downsample} and {\myfont Sparse} into {\myfont Hybrid}. The DSI contains a soft estimation of the disparities, and we wish to utilize information from both approaches at this stage, prior to the hard decision (depth selection).

The algorithm is depicted in \ref{fig:Flowcharts}(c). Let $\bar{I}_2$ denote a sparse sampling of $I_2$.  We compute a weighted mean of the estimated DSI $\hat{{\cal D}}$ and a second DSI, calculated from interpolation of the samples in $\bar{I}_2$. This second DSI can be seen in the bottom route, surrounded by the dash-dot blue line. A direct interpolation of the samples would introduce aliasing, so we apply a low pass filter on the interpolated image, according to the transmitted frequencies. We also smooth $I_1$ to match the frequencies, just like we did in {\myfont Downsample}. The blocks surrounded by the dotted red line are identical to {\myfont Sparse}, while the route surrounded by the dash-dot blue line is inspired by {\myfont Downsample} (the difference stems from the different input $\bar{I}_2$).

Given the interpolated $\bar{I}_2$, the left-right consistency check such as the one described in \cite{cvpr-12-qingxiong-yang} can be performed. It significantly improves the disparity map's accuracy.

\subsection{A lower bound}\label{LowerBound}

We report results of the encoding schemes in the experimental section, but as with any encoding scheme one wonders: Are we making the most out of the bandwidth at our disposal? How far are we from optimal encoding? We measure the optimal encoding as follows.

Let ${\cal D}$ be the full DSI computed from the {\em original} $I_1$ and $I_2$. This is the best we can hope for. Now, resize ${\cal D}$ down to the size of the allocated bandwidth. If we take image resize to be the optimal encoding of a signal, then this gives the optimal encoding. In the experimental section we use this technique to measure how far are the encoding schemes from optimal encoding.


\subsection{Recovering $I_2$}\label{RecoveringI2}
So far we discussed recovering depth. Given $I_1$ and the recovered depth, $\hat{I_2}$ (an estimate of $I_2$) can be synthesized using depth-image based rendering techniques. This is beyond the scope of this paper, but we will describe briefly the method we used. Recovering $I_2$ involves three major steps: \emph{warp}, \emph{inpainting} and \emph{enhancement}.

Given $I_1$ and $D$, it is possible to \emph{warp} $I_1$ and obtain $\hat{I_2}$, the scene as viewed from the position of the camera who produced $I_2$, minus occluded areas (see \cite{Szeliski-Scharstein} for details).

Although the pixels in the occluded areas can't be retrieved from $I_1$, we have additional information in $\bar{I}_2$. We use this information to \emph{inpaint} the occluded areas. In case of {\myfont Downsample}, we can copy the missing pixels from the upsampled $\bar{I}_2$. In case $\bar{I}_2$ contains sparse samples from $I_2$ ({\myfont Sparse} and {\myfont Hybrid}), we can interpolate the samples (after smoothing) and use them to fill the occluded areas. If the occlusion is small, or if it doesn't contain texture, the blur will not be significant.

Furthermore, in the cases of {\myfont Sparse} and {\myfont Hybrid}, $\bar{I}_2$ contains exact samples of $I_2$.
We use those samples to \emph{enhance} $\hat{I_2}$ in the non-occluded areas as follows: we calculate a sparse difference image between $\bar{I}_2$ and $\hat{I_2}$ (it's sparse because $\bar{I}_2$ is sparse).
We then perform joint bilateral filtering on the difference image with $\hat{I_2}$ as the guidance image, and finally we add the filtered difference image to $\hat{I_2}$ in the non-occluded areas.

\section{Connection to Information Theory}

Our problem has its roots in the Distributed Source Coding literature. Consider two sources $I_1$ and $I_2$ that are known to be correlated. Suppose both $I_1$ and $I_2$ are known at the encoder but only $I_1$ is known at the decoder, and assume the encoder wishes to efficiently transmit $I_2$ to the decoder. Clearly the encoder can take advantage of the fact that $I_1$ is known to both sides to better encode $I_2$. The remarkable result of Slepian and Wolf \cite{Slepian-Wolf} is that the encoder can encode $I_2$ just as well {\em without} knowing $I_1$ at all. The basic result of Slepian and Wolf holds for lossless compression and it was later extended by Wyner and Ziv \cite{Wyner-Ziv} to the lossy case.

These results were theoretical and the first practical implementation was reported in DISCUS \cite{DISCUS}. This led to a surge of interest in developing efficient Wyner-Ziv video encoding algorithms. See \cite{Girod-invited-paper,Pereira-Brites-DVC-survey,DVC-survey} for a review of recent advances in the field.
-

\section{Results}

We first evaluate the performance of the different algorithms by measuring the accuracy of the initial disparity map. We evaluate the result on the Middlebury benchmark stereo datasets \cite{Szeliski-Scharstein,teddy-cones-cite}. We also show how our algorithm compares with a standard compression technique such as JPEG2000. Next, we compare our recovered images to a distributed video codec. We use the recovery process $\hat{I_2}$ described in \ref{RecoveringI2} in order to make this comparison.
Finally we show results of our algorithm on indoor and outdoor scenes. Since the true depth is not available for these scenes, we measure results using the recovered $\hat{I_2}$.


\subsection{Stereo}

\begin{figure*}[tb]
\begin{center}
\begin{tabular}{cc}
\includegraphics[width=0.43\linewidth]{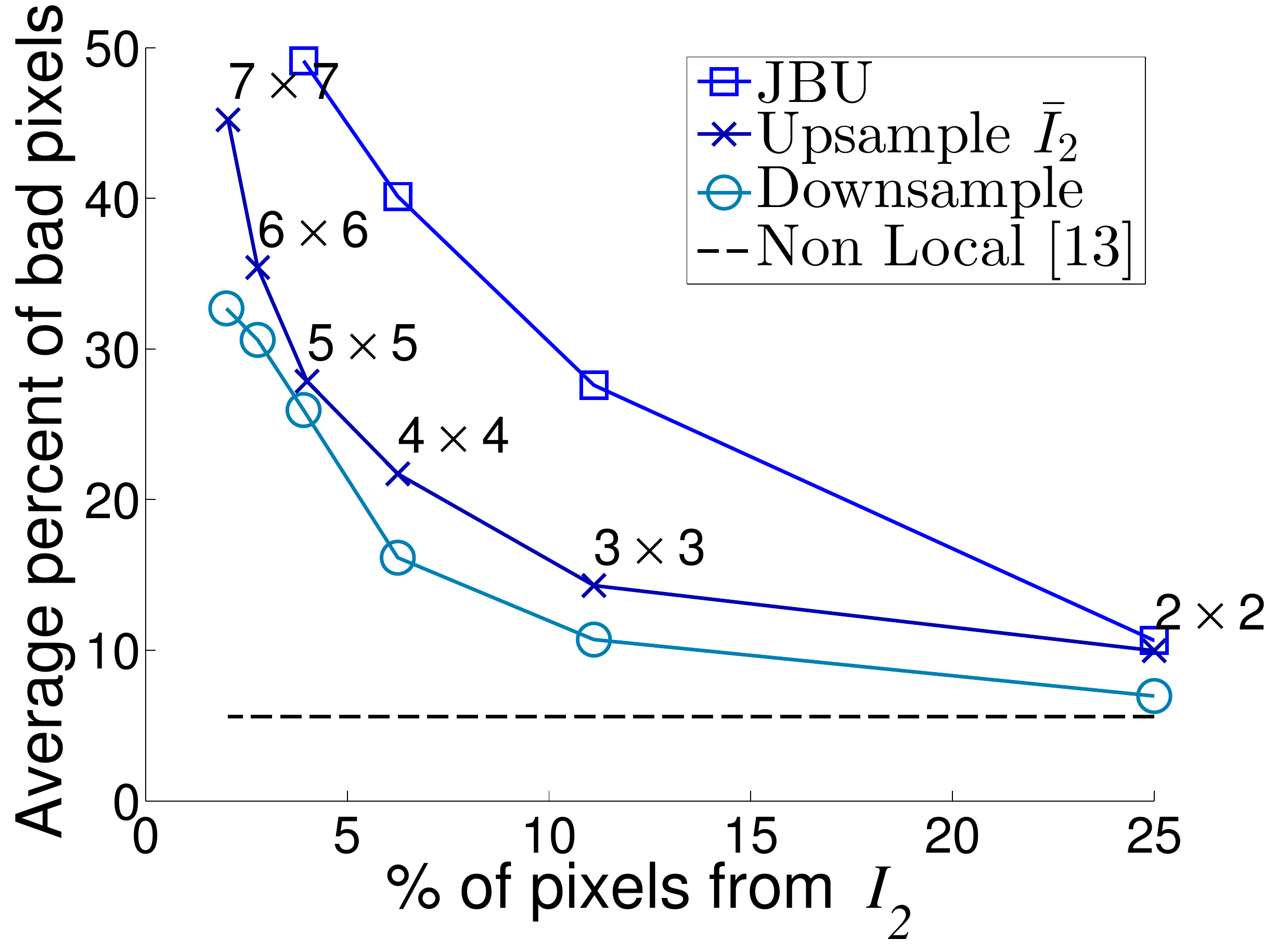} &
\includegraphics[width=0.43\linewidth]{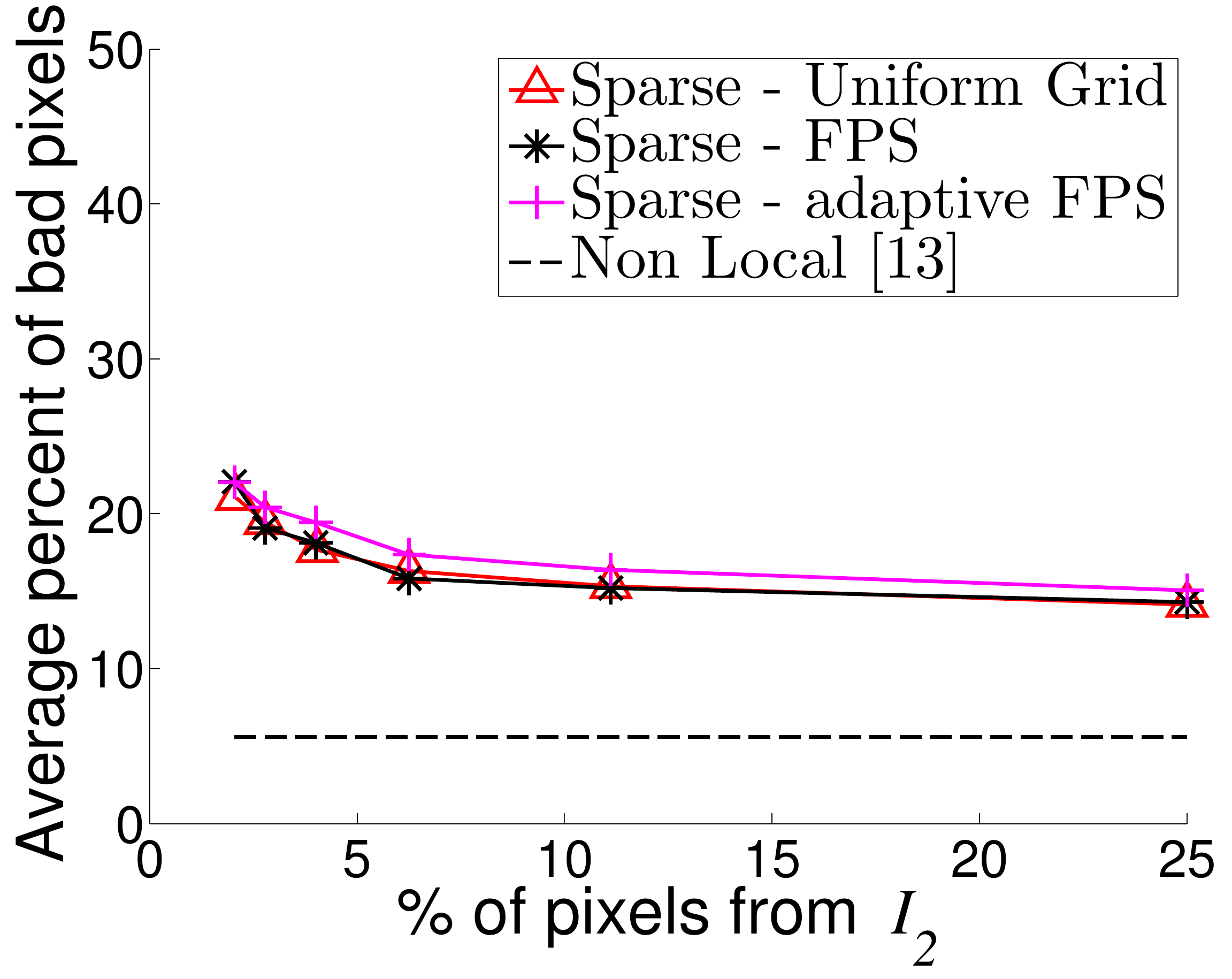} \\
 (a) Downsample & (b) Sparse \\
\includegraphics[width=0.43\linewidth]{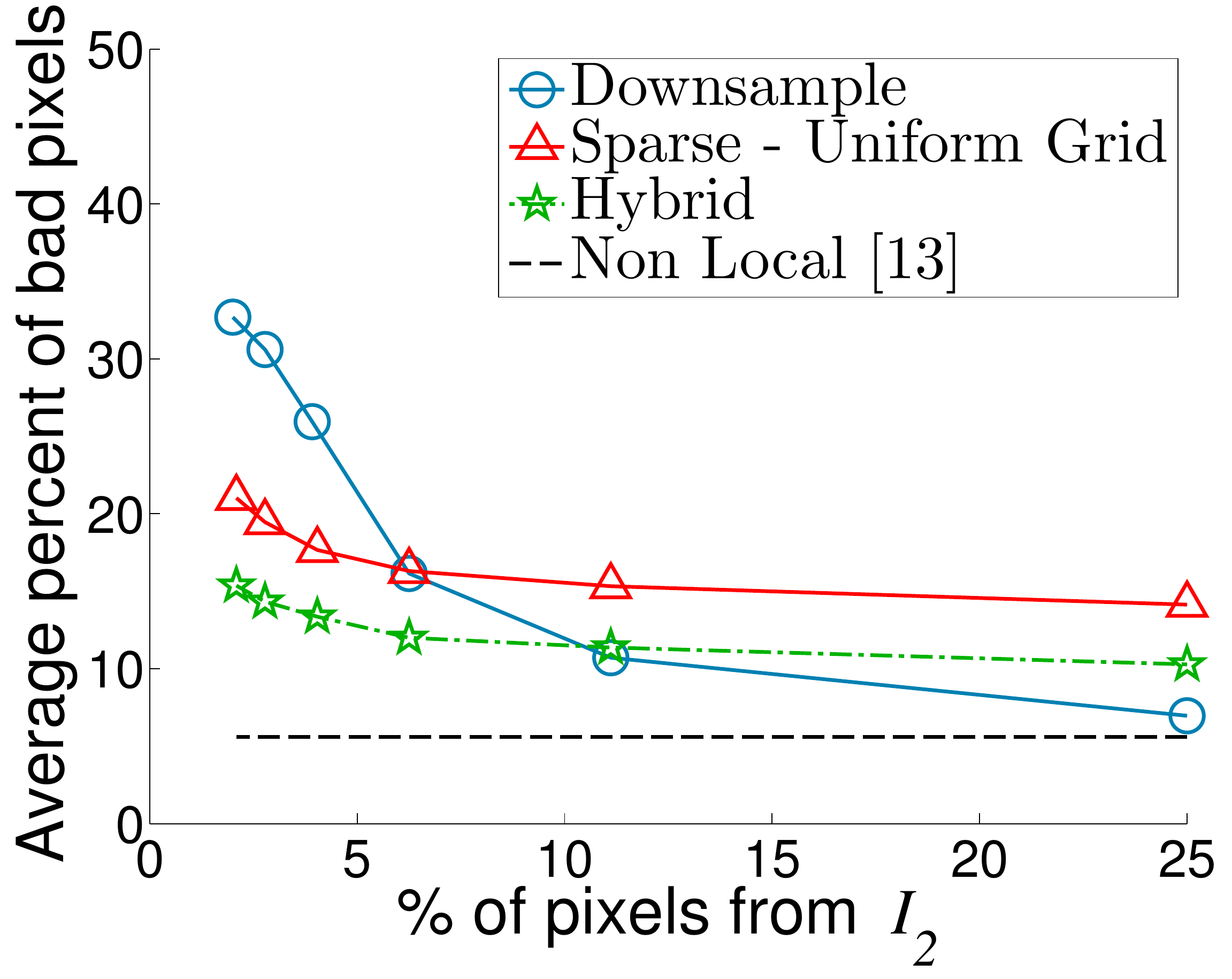} &
\includegraphics[width=0.43\linewidth]{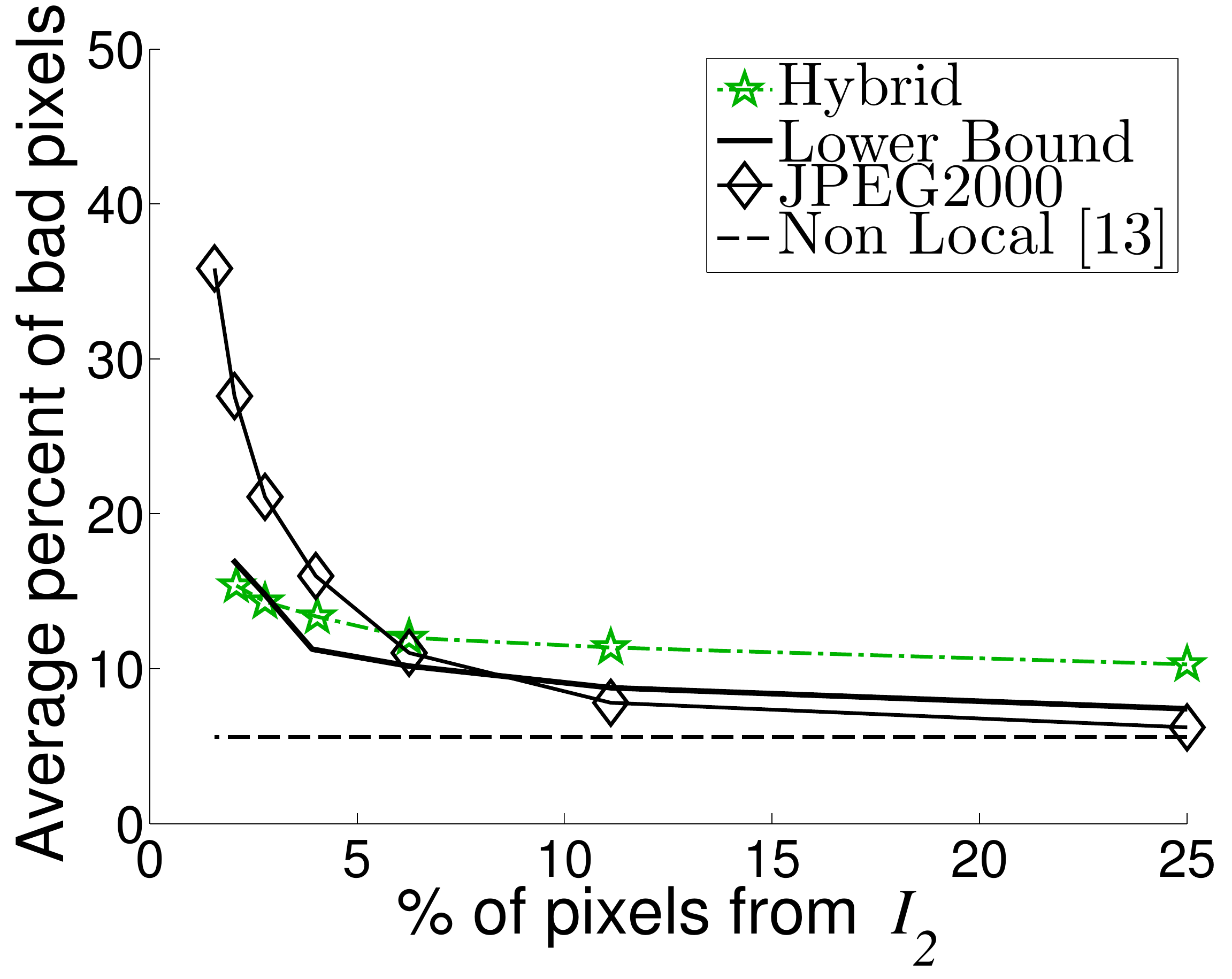} \\
 (c) Downsample, Sparse, Hybrid  & (d)  Sparse vs. JPEG2000 \\
\end{tabular}
\end{center}
\caption{Average performance of different "stereo on a budget" strategies on the middlebury stereo dataset. The x-axis is the percentage of $I_2$ used for the calculation of the disparity map, the y-axis is the average percent of pixels with disparity error larger than 1. When a standard stereo algorithm is used, we use \cite{cvpr-12-qingxiong-yang}. Its performance is included for reference (using 100\% of $I_2$ for the calculation). The labels on the graph denote the scaling factor of the image, in both axes.}
\label{fig:CompareAll}
\end{figure*}

\begin{figure*}[tb]
\begin{center}
\begin{tabular}{cccc}
\includegraphics[width=0.22\linewidth]{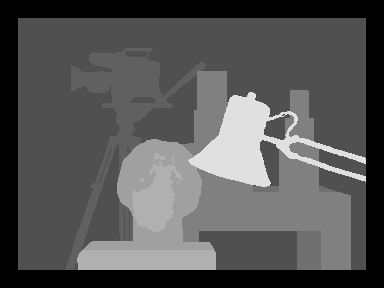} &
\includegraphics[width=0.22\linewidth]{tsukuba_imresize_both_5_5_LRconsistent.png} &
\includegraphics[width=0.22\linewidth]{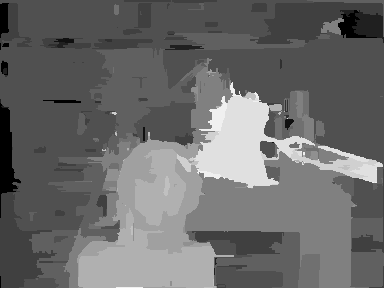} &
\includegraphics[width=0.22\linewidth]{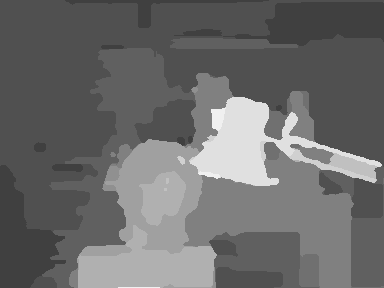} \\
\includegraphics[width=0.22\linewidth]{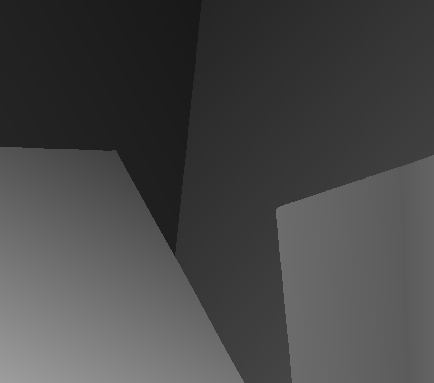} &
\includegraphics[width=0.22\linewidth]{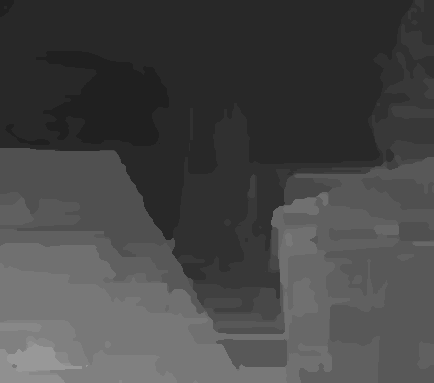} &
\includegraphics[width=0.22\linewidth]{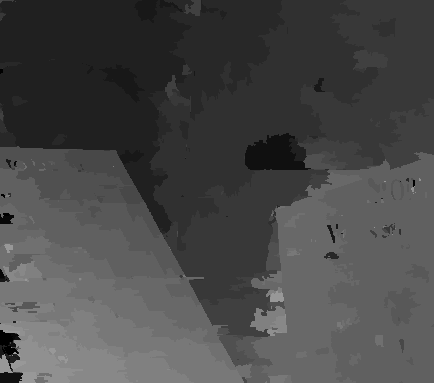} &
\includegraphics[width=0.22\linewidth]{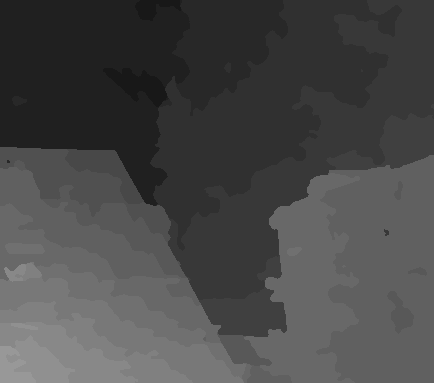} \\
\includegraphics[width=0.22\linewidth]{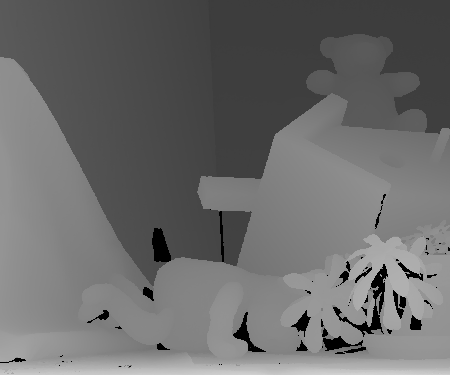} &
\includegraphics[width=0.22\linewidth]{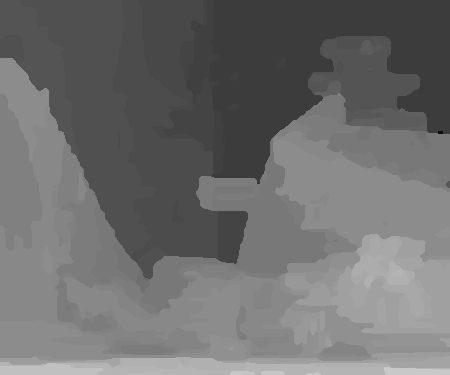} &
\includegraphics[width=0.22\linewidth]{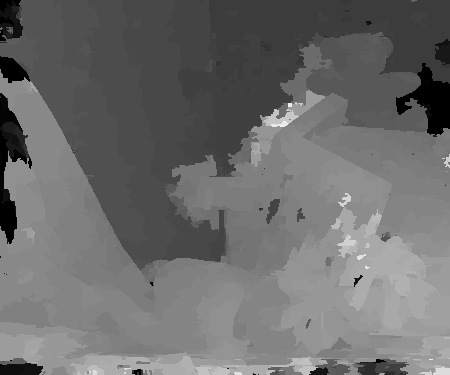} &
\includegraphics[width=0.22\linewidth]{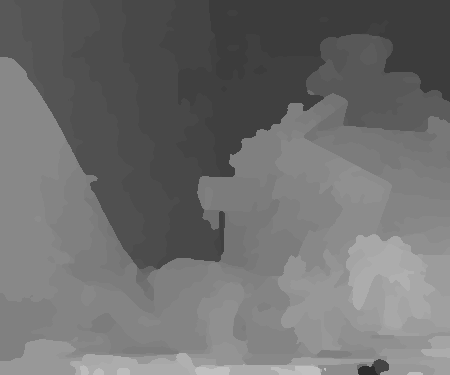} \\
\includegraphics[width=0.22\linewidth]{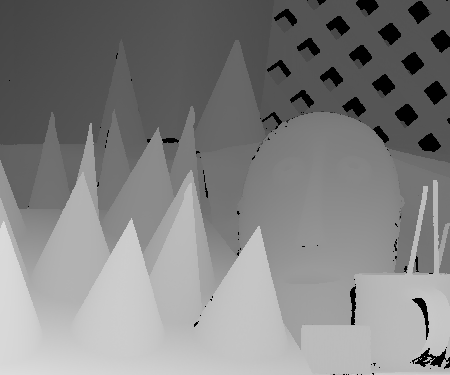} &
\includegraphics[width=0.22\linewidth]{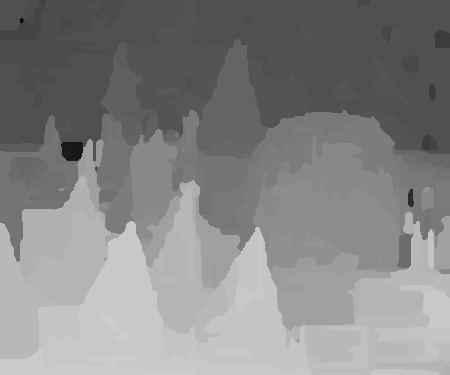} &
\includegraphics[width=0.22\linewidth]{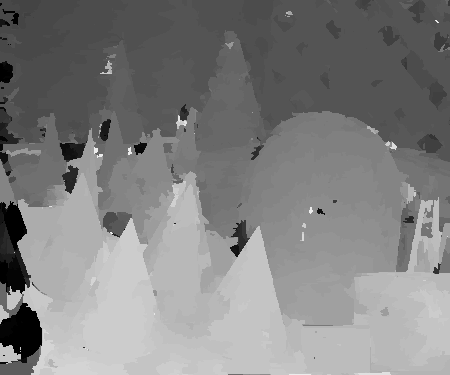} &
\includegraphics[width=0.22\linewidth]{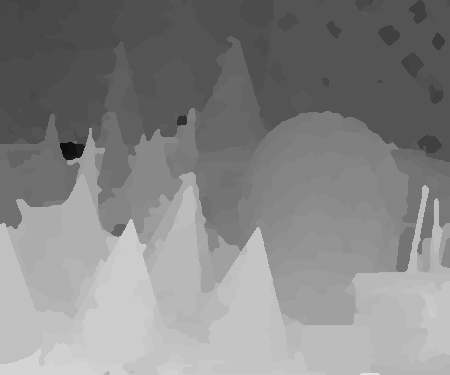}  \\
groundtruth &   {\myfont Downsample} &   {{\myfont Sparse} - uniform grid} &   {\myfont Hybrid} \\
\end{tabular}
\end{center}
\caption{Results for Middlebury benchmark data. The first column shows the groundtruth disparity maps.
The other columns are disparity maps calculated using only 4\% of $I_2$.}
\label{fig:MiddleburyDisparityMaps}
\end{figure*}

First, we evaluate the performance of {\myfont Downsample} vs. other approaches, using as input a downsampled version of $I_2$. Figure \ref{fig:CompareAll}(a) shows that {\myfont Downsample} as described in Figure \ref{fig:Flowcharts}(a) ranks higher than computing a low resolution disparity and performing joint bilateral upsampling (denoted JBU) as suggested in \cite{joint-bilateral-upsampling}. In addition, the disparity map is more accurate in terms of number of bad pixels when $I_1$ is smoothed compared to when it's not smoothed (denoted "Upsample $\bar{I}_2$").

When the images are resized to half of their original dimension there is very little loss of quality compared to no-resizing at all, since the correlation between adjacent pixels is high. When the image is further resized, the quality of the disparity map drops. We are interested in sending only a small fraction of the data, which would require to downsize $I_2$ by at least 5 in each dimension. Computing disparity in low-resolution and JBU would require an extremely good sub-pixel accuracy, which is rare for large scaling factors. Hence we compute the stereo matching in full resolution.

Next, we compare different sparse sampling strategies. The uniform sampling over a grid is more appealing in terms of computation power, however it may be sub-optimal in terms of aliasing. Therefore, we compare it's performance to FPS - random uniform sampling \cite{FarthestPointSampling}, and to adaptive-FPS, random sampling which is part uniform (80\%) and part adaptive (last 20\% of the samples). The disparity maps were calculated according to {\myfont Sparse}, as described in Figure \ref{fig:Flowcharts}(b).
Figure \ref{fig:CompareAll}(b) shows that in terms of percentage of bad pixels in the disparity map, there isn't a significant difference between the different sampling methods. Also in terms of RMS error the sampling strategies are quite comparable, as evident in tables \ref{MiddelburySparseRMSError} and \ref{MiddelburySparseBadPixels}. Therefore, the uniform grid sampling is preferable.

For the Joint-Bilateral filter, we used the parameters $\sigma_r=20$ and $\sigma_s=3 \cdot (grid\ spacing)$ in Equation \ref{eq:sparse-dsi-filtering}, where $(grid\ spacing)$ is the sub-sampling factor in the {\myfont Sparse} algorithm (see \cite{Chaudhury:2011:FOB} for an explicit definition of the filter).

\begin{table*}[!t]
\centering
\renewcommand{\arraystretch}{1.3}
\caption{RMS error of the disparity maps, calculated using only 4\% of $I_2$. In the adaptive FPS, we sampled 80\% of the samples uniformly, and only the last 20\% of the samples were adaptive. Otherwise large region of the image wouldn't be covered, which would lead to errors.}
\label{MiddelburySparseRMSError}
\begin{tabular}{|l|ccc|ccc|ccc|ccc|}
\hline
Algorithm & \multicolumn{3}{c|}{{\em Tsukuba}} & \multicolumn{3}{c|}{{\em Venus}} & \multicolumn{3}{c|}{{\em Cones}} &  \multicolumn{3}{c|}{{\em Teddy}} \\
             & nonocc & all  & disc & nonocc & all  & disc & nonocc & all  & disc & nonocc & all  & disc \\
 \hline \hline
Uniform grid &  1.33  &	1.45 & 2.43 &  1.28  & 1.49	& 1.74 &  5.41	& 8.56 & 8.62 &	 4.33  & 9.45 &	 7.41 \\
FPS          &  1.37  & 1.52 & 2.69 &  1.23  & 1.48	& 1.61 &  5.05  & 8.29 & 7.31 &	 4.68  & 9.62 &  8.01 \\
Adaptive FPS &  1.36  & 1.49 & 2.49	&  1.57	 & 1.77	& 2.14 &  5.75  & 8.33 & 8.12 &  4.60  & 9.75 &	 7.78 \\
\hline
\end{tabular}
\end{table*}

\begin{table*}[!t]
\centering
\renewcommand{\arraystretch}{1.3}
\caption{Bad pixels (with error $ > 1$) in the disparity maps, calculated using only 4\% of $I_2$. In the adaptive FPS, we sampled 80\% of the samples uniformly, and only the last 20\% of the samples were adaptive. Otherwise large region of the image wouldn't be covered, which would lead to errors.}
\label{MiddelburySparseBadPixels}
\begin{tabular}{|l|ccc|ccc|ccc|ccc|}
\hline
Algorithm & \multicolumn{3}{c|}{{\em Tsukuba}} & \multicolumn{3}{c|}{{\em Venus}} & \multicolumn{3}{c|}{{\em Cones}} &  \multicolumn{3}{c|}{{\em Teddy}} \\
             & nonocc & all  & disc & nonocc & all  & disc & nonocc & all  & disc & nonocc & all  & disc \\
 \hline \hline
Uniform grid &  7.30  & 8.86 & 20.55& 10.03	 &11.12 &19.79 & 20.37	& 26.66& 38.61&	 14.01 &24.04 & 29.65 \\
FPS          &  6.61  & 8.18 & 24.50& 10.49	 &11.62	&21.39 & 20.91  & 28.22& 36.58&  14.49 &24.47 & 29.59 \\
Adaptive FPS &  7.49  & 9.20 & 24.50& 11.93	 &12.96	&25.16 & 24.01  & 30.43& 38.81&  15.44 &25.17 & 31.87 \\
\hline
\end{tabular}
\end{table*}

Figure \ref{fig:CompareAll}(c) compares {\myfont Sparse}, {\myfont Downsample} and {\myfont Hybrid} strategies (Figure \ref{fig:Flowcharts}). For {\myfont Sparse} we used uniform grid sampling, since it is the most efficient in terms of computation and power savings. At strong compression ratios, {\myfont Hybrid} is superior to using either strategy alone.
Noteworthy is the fact that with only 11.1\% of $I_2$ we can compute a disparity map with an accuracy which is comparable to that of graph-cut on $I_1$ and $I_2$.

Figure \ref{fig:CompareAll}(d) compares {\myfont Hybrid} and standard JPEG2000 compression with variable rate on $I_2$. Noteworthy is the fact that at extreme compression regimes, retrieving depth from $\bar{I}_2$ and $I_1$ using {\myfont Hybrid} is more accurate than using $I_1$ and a JPEG2000 compressed $I_2$. Also shown is the lower bound estimation as described in section \ref{LowerBound}.

Some of the disparity maps can be seen in Figure \ref{fig:MiddleburyDisparityMaps}, using only 4\% of $I_2$ for the calculation.
In {\myfont Downsample} the high frequencies are lost in the images, and hence also in the depth map, \eg the missing lamp arm in tsukuba. In {\myfont Sparse} the fine details are better preserved, such as the shape of the cones and the video camera in tsukuba. {\myfont Hybrid} combines the best of both, and we will focus on that algorithm.

\begin{figure*}[tb]
\begin{centering}
\small\addtolength{\tabcolsep}{-3pt}
\begin{tabular}{ccccc}
{\small Original $I_2$} & {\small $I_1$ warped to $I_2$} & {\small $I_1$ warped to $I_2$} & {\small $I_1$ warped to $I_2$}  & {\small warp with}\\
{\small }               & {\small Sparse 11.1\% pixels}  & {\small Sparse 11.1\% pixels}  & {\small Sparse 11.1\% pixels}   & {\small Sparse 4\% pixels}\\
{\small  }              & {\small  }                     & {\small + enhancement }        & {\small + inpainting}           & {\small + enhancement, inpainting}\\
\includegraphics[width=0.185\linewidth]{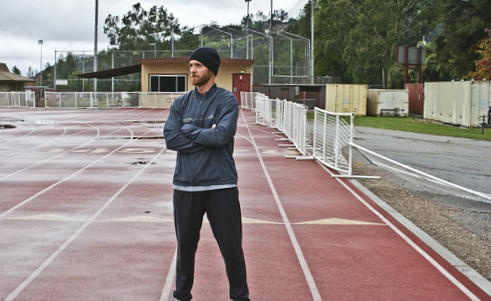} &

\includegraphics[width=0.185\linewidth]{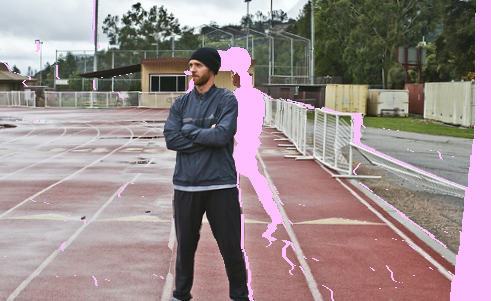} &
\includegraphics[width=0.185\linewidth]{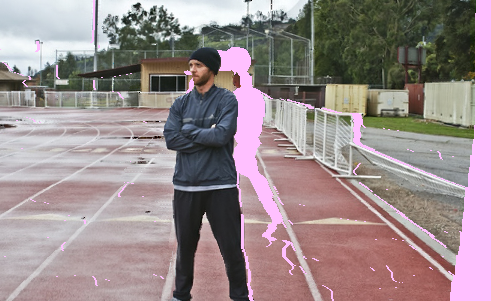} &
\includegraphics[width=0.185\linewidth]{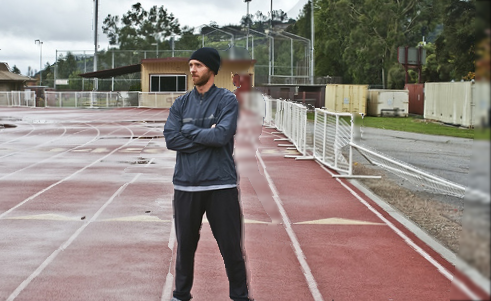} &
\includegraphics[width=0.185\linewidth]{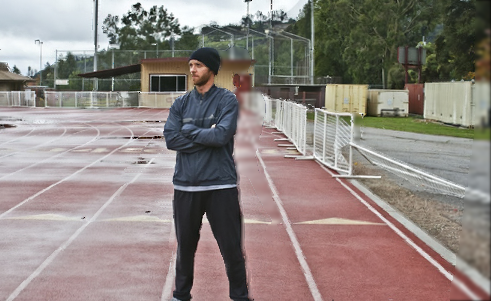} \\
{\small } & {\small PSNR: 27.17 dB} & {\small PSNR: 27.66 dB} & {\small PSNR: 26.24 dB} & {\small PSNR: 26.60 dB} \\
\end{tabular}
\end{centering}
\caption{Demonstration of the estimation process of $\hat{I_2}$ described in section \ref{RecoveringI2} on an outdoor scene. This image was downloaded from flickr and manually rectified by \cite{Basha:2011:GCS}.
The first column shows $I_2$, the right image of the stereo pair.
The 2\textsuperscript{nd} column shows the warp results using the disparity map calculated using the {\myfont Sparse} method, with $\bar{I_2}$ containing 11.1\% of the pixels in $I_2$. The PSNR is calculated on the non-occluded areas.
The 3\textsuperscript{rd} column is after the enhancement step described in \ref{RecoveringI2}, which gives increases the PSNR by about 0.5dB.
The 4\textsuperscript{th} column is the 2\textsuperscript{nd} column with the occluded areas now inpainted with interpolation of the sample from $\bar{I_2}$.
The 5\textsuperscript{th} column is the inpainting result of the image from the 3\textsuperscript{rd} column.}
\label{fig:RecoverI2Example}
\end{figure*}

Figure \ref{fig:RecoverI2Example} demonstrates the process of recovering $I_2$, which is described in section \ref{RecoveringI2}. $I_2$ is encoded in-camera and is estimated at the host by {\em warping} $I_1$ with the calculated disparity map and using the available samples from $\bar{I_2}$ to {\em enhance} the result and {\em inpaint} the occluded areas.

Figure \ref{fig:RateDistortionI2} shows the effectiveness of the enhancement process described in section \ref{RecoveringI2}, showing the PSNR before (dashed lines) and after the enhancement (solid lines), for the Middlebury datasets. Once an estimate of $I_2$ is calculated, it can be used as input to other algorithms. Thus future advancements in stereo matching as well as different types of algorithms can benefit from our algorithm.

\subsection{Comparison to Distributed Video Coding}\label{DVCcomparison}

In \cite{VarodayanStereo} Varodayan \etal developed a coding scheme that exploits the similarity of stereo images without communication among the cameras. It was later extended to video \cite{VarodayanVideo}, and the code is available online.

The code was designed as a video codec, and its output is the reconstructed frame, not the disparity map. Because it is a video codec, it works for general two frames and does not take advantage of the fact that the images are rectified. Therefore, the motion search space is two-dimensional and is limited to a motion field of 5 pixels in each direction (the number of options totals $11 \times 11$ possibilities), even though the true disparity is larger.
The code only accepts images in QCIF resolution, so we downsampled the standard Middlebury datasets (and cropped if necessary, to maintain the aspect ratio). The rate is determined automatically: the decoder may request more bits of information from the encoder via a feedback channel if the reconstructed image isn't good enough. Our algorithm does not use a feedback channel. The running time of DVC is several minutes, while ours is an optimized Matlab code that takes a few seconds to run.

\begin{figure}[tb]
\begin{center}
\begin{tabular}{cr}
\includegraphics[width=0.4\linewidth]{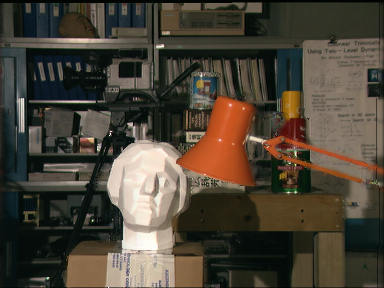} &
\includegraphics[width=0.4\linewidth]{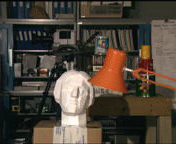} \\
PSNR: 29.64 dB & PSNR: 30.76dB, 12.6\% \\
\includegraphics[width=0.35\linewidth]{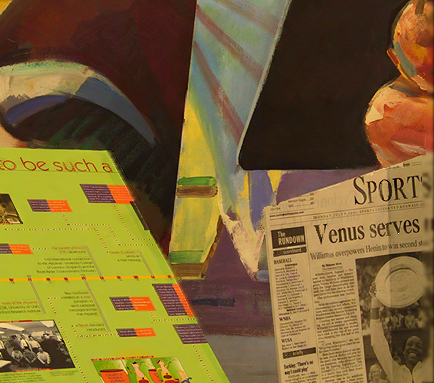} &
\includegraphics[width=0.35\linewidth]{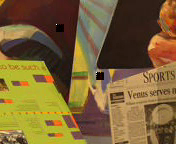} \\
PSNR: 30.23 dB & PSNR: 26.86dB, 15.21\% \\
\includegraphics[width=0.4\linewidth]{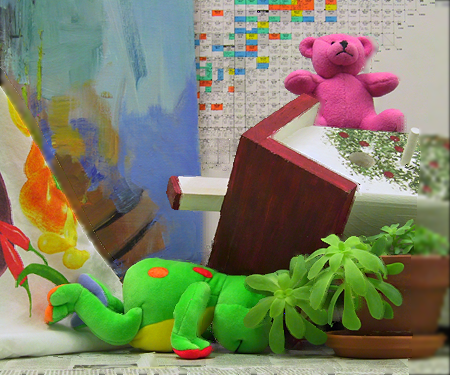} &
\includegraphics[width=0.4\linewidth]{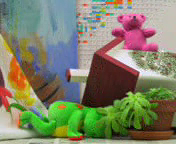} \\
PSNR: 27.45 dB & PSNR: 25.6dB, 22.96\% \\
\includegraphics[width=0.35\linewidth]{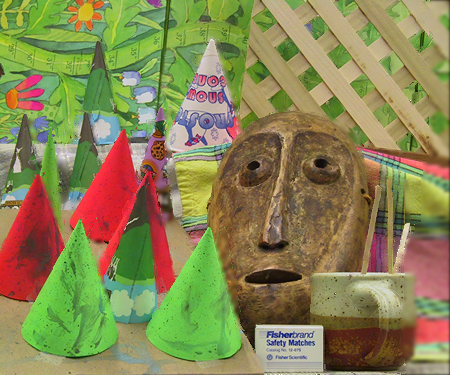} &
\includegraphics[width=0.35\linewidth]{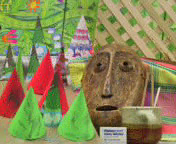} \\
PSNR: 26.48 dB & PSNR: 25.01dB, 27.96\%  \\
\end{tabular}
\end{center}
\caption{Comparing our method (left column) to the DVC method of \cite{VarodayanVideo} (right column). We show recovered and enhanced $\hat{I_2}$ using our method with only 11.1\% of $I_2$ used to calculate the disparity map. The DVC method determines the percentage of transmitted pixels adaptively. We give these numbers below each image. (See more results in supplemental material).}
\label{fig:VarodayanResults}
\end{figure}

\begin{figure}
\begin{center}
\begin{tabular}{cc}
\includegraphics[width=0.4\linewidth]{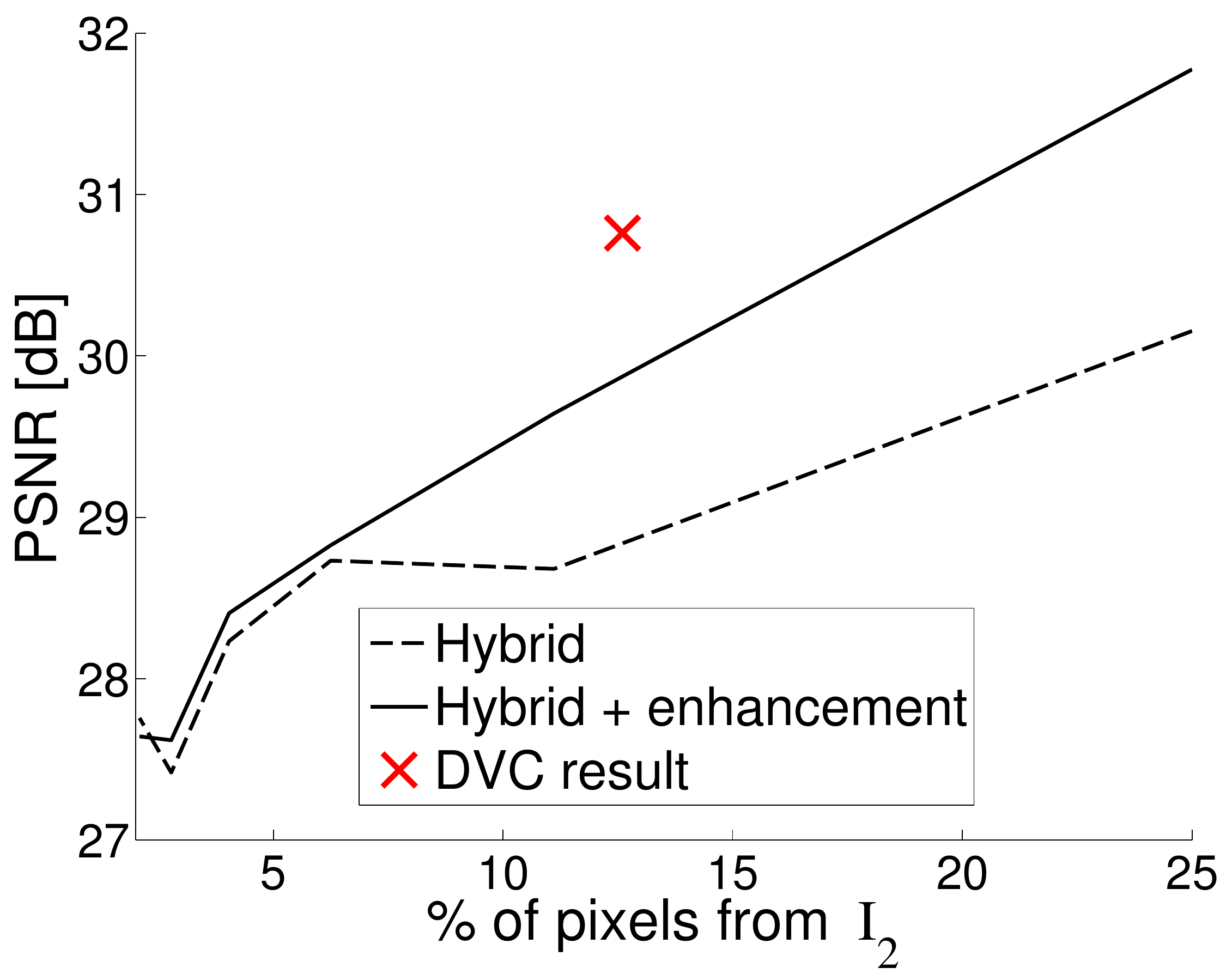} &
\includegraphics[width=0.4\linewidth]{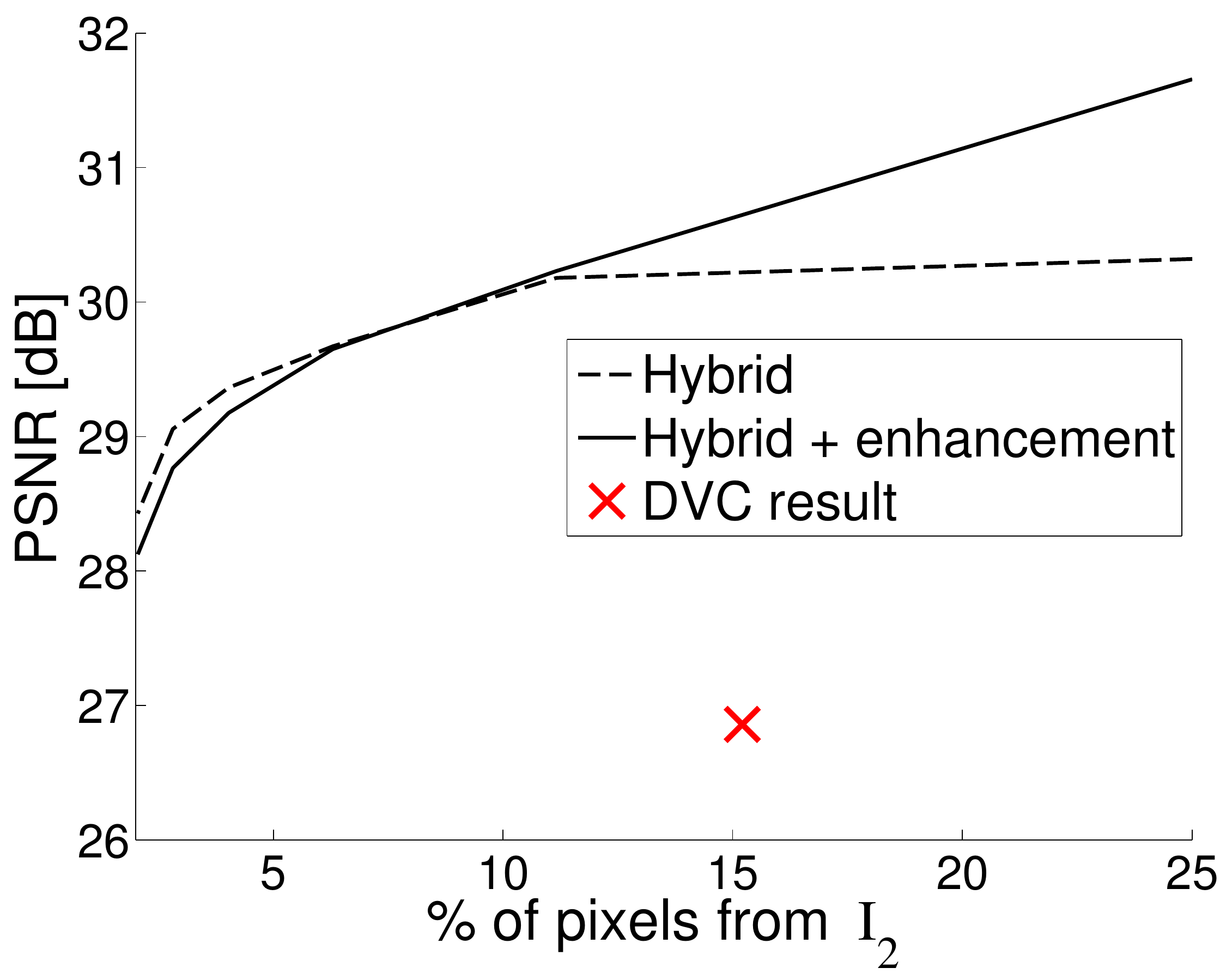} \\
tsukuba & venus \\
\includegraphics[width=0.4\linewidth]{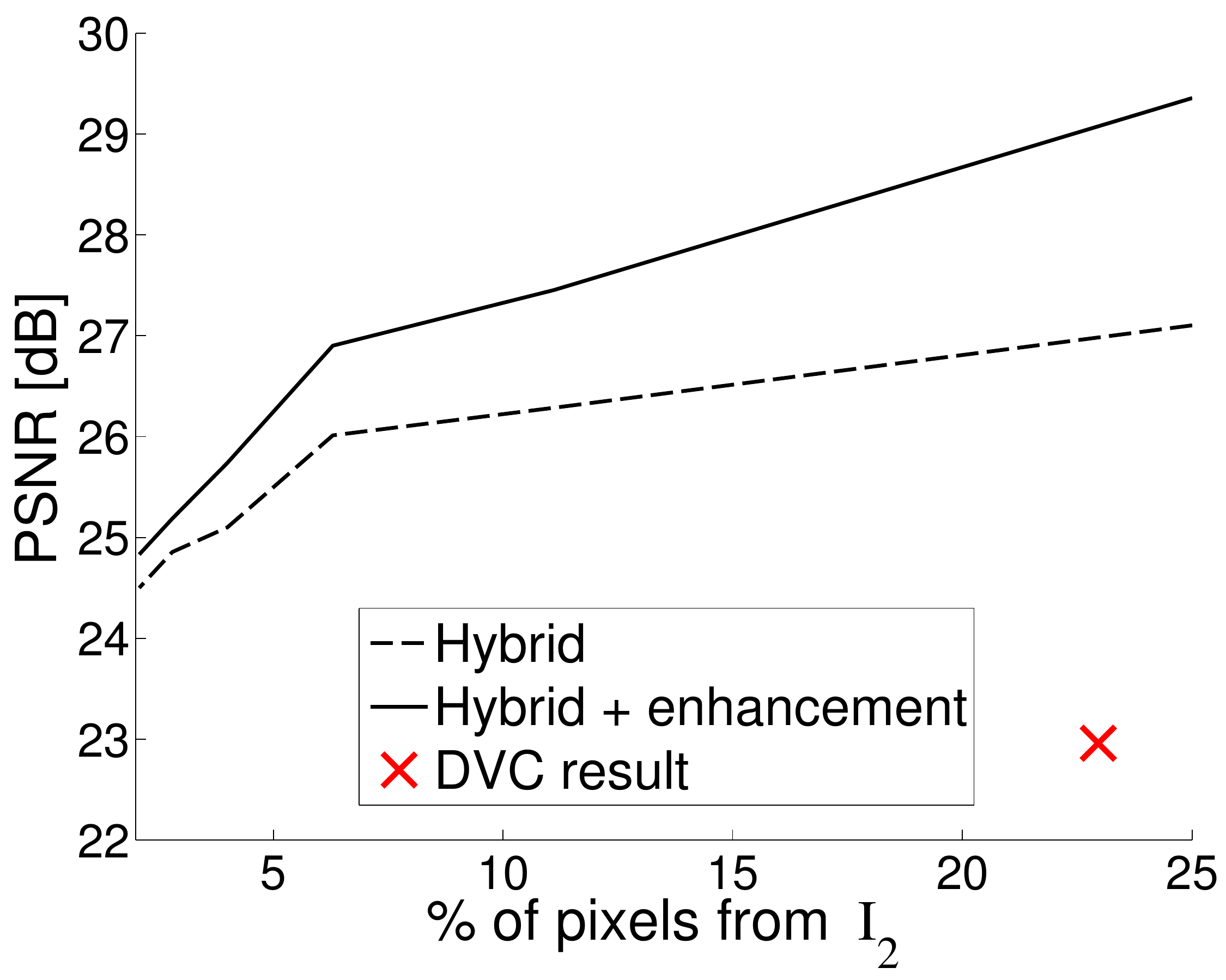} &
\includegraphics[width=0.4\linewidth]{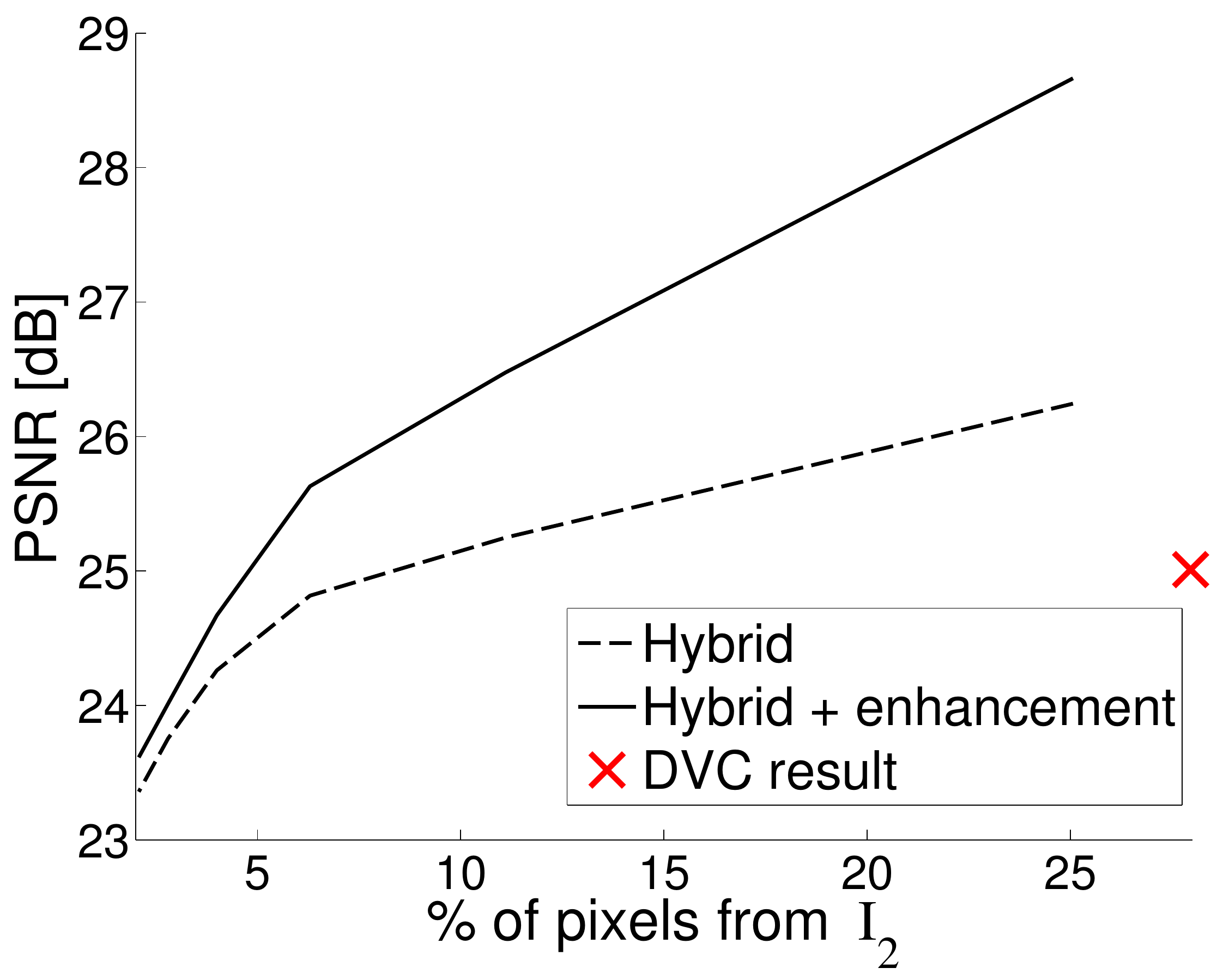} \\
teddy & cones \\
\end{tabular}
\end{center}
\caption{Rate distortion curves of $\hat{I_2}$ recovered using the {\myfont Hybrid} algorithm, for the Middluebury datasets. The graph shows the PSNR before (dashed lines) and after (solid lines) the enhancement, showing a larger gain when more samples of $I_2$ are given. For comparison, the results obtained from the code of \cite{VarodayanVideo} are shown on the graphs.}
\label{fig:RateDistortionI2}
\end{figure}

Figure \ref{fig:VarodayanResults} shows both our results and \cite{VarodayanVideo}'s. On average we achieve higher PSNRs, while transmitting a smaller fraction of the image, and the computation time is shorter than the DVC codec. The DVC results from Figure \ref{fig:VarodayanResults} can be compared to the rate-distortion curves in Figure \ref{fig:RateDistortionI2}.

\begin{figure}[!t]
\begin{centering}
\small\addtolength{\tabcolsep}{-2pt}
\begin{tabular}{ccc}
  Original $I_2$ &   enhanced $\hat{I_2}$, 11.1\% &    enhanced $\hat{I_2}$, 4\%\\
\includegraphics[width=0.31\linewidth]{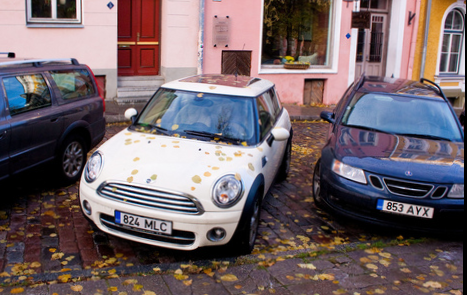} &
\includegraphics[width=0.31\linewidth]{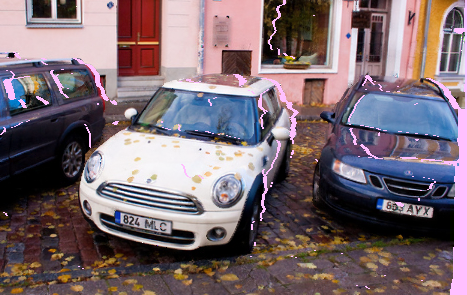} &
\includegraphics[width=0.31\linewidth]{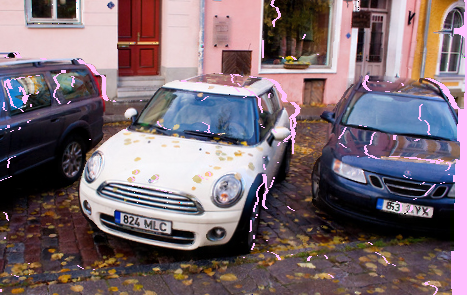} \\
   &   PSNR: 27.02 dB &    PSNR: 25.4 dB\\
\includegraphics[width=0.31\linewidth]{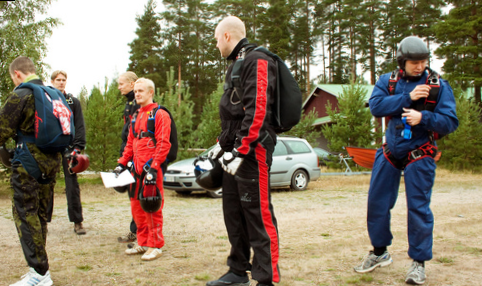} &
\includegraphics[width=0.31\linewidth]{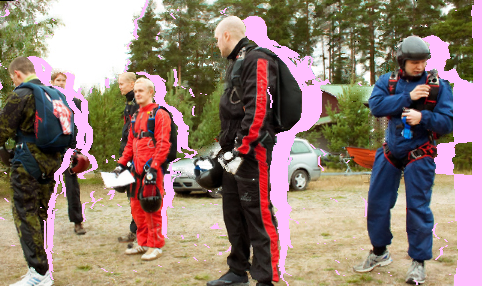} &
\includegraphics[width=0.31\linewidth]{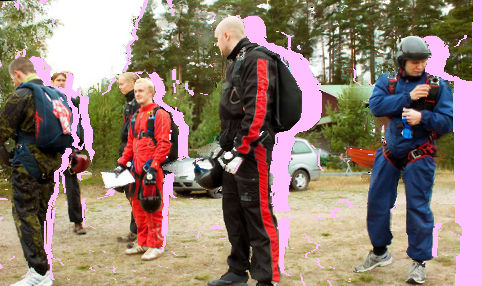} \\
   &   PSNR: 23.4 dB &   PSNR: 21.7 dB\\
\includegraphics[width=0.31\linewidth]{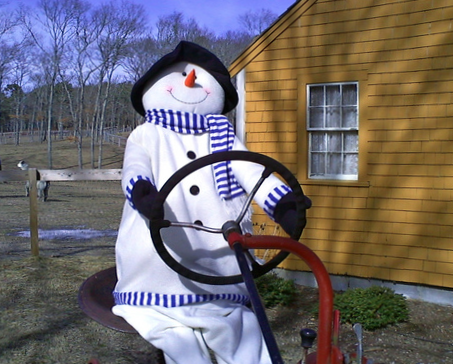} &
\includegraphics[width=0.31\linewidth]{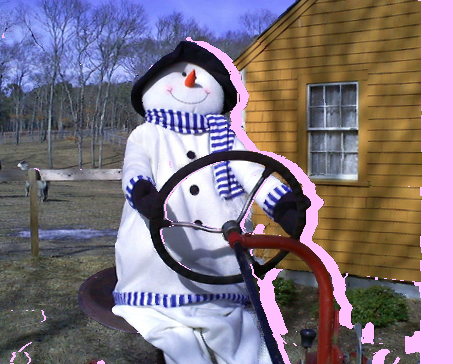} &
\includegraphics[width=0.31\linewidth]{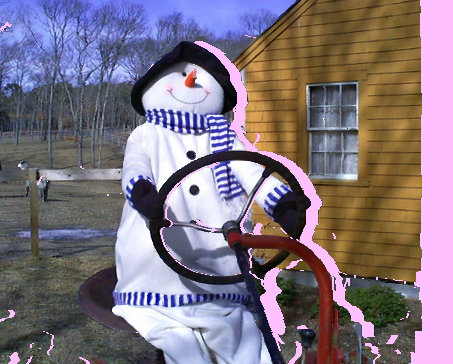} \\
   &   PSNR: 27.8 dB &    PSNR: 26.8 dB\\
\includegraphics[width=0.31\linewidth]{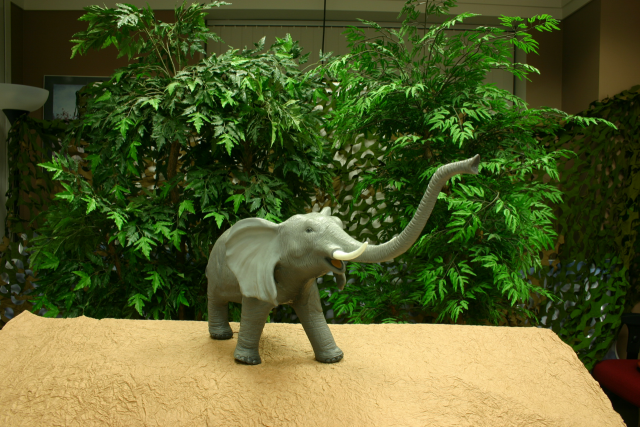} &
\includegraphics[width=0.31\linewidth]{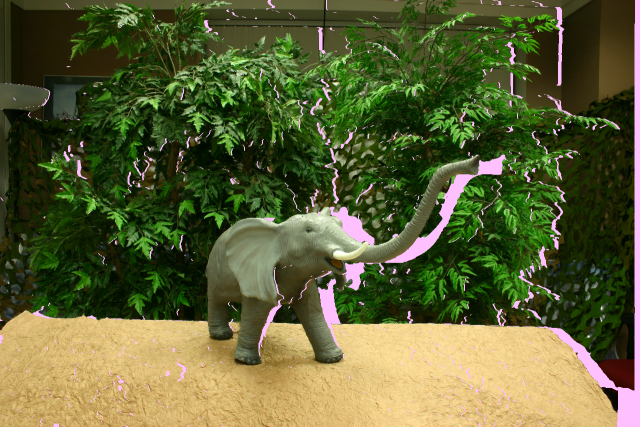} &
\includegraphics[width=0.31\linewidth]{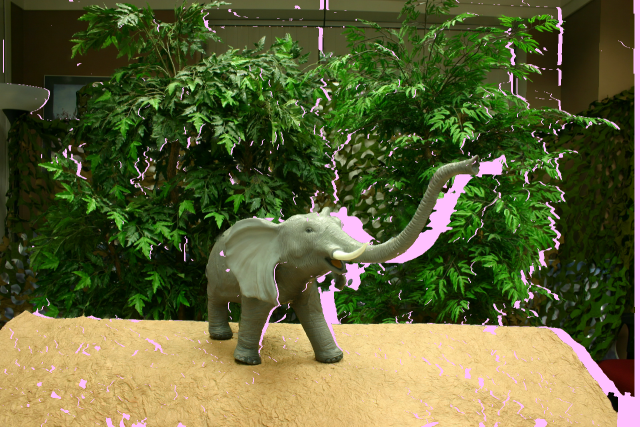} \\
   &    PSNR: 28.94 dB &   PSNR: 27.3 dB\\
\includegraphics[width=0.31\linewidth]{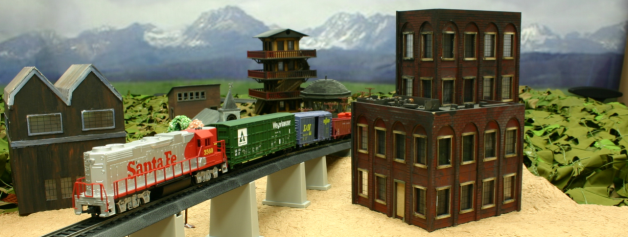} &
\includegraphics[width=0.31\linewidth]{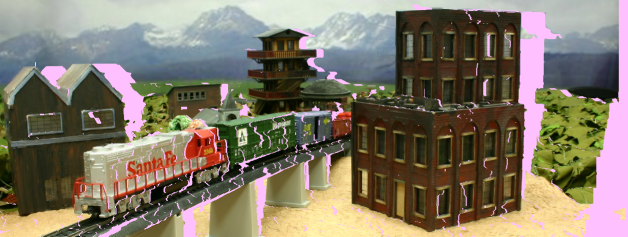} &
\includegraphics[width=0.31\linewidth]{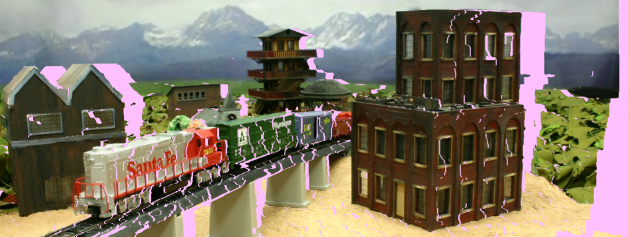} \\
   &   PSNR: 30.2 dB &   PSNR: 27.95 dB\\
\end{tabular}
\end{centering}
\caption{Results of our algorithm on various stereo pairs (the stereo images can be found in the supplementary material). The first column is $I_2$; the 2\textsuperscript{nd} and 3\textsuperscript{rd} columns show the enhanced $\bar{I_2}$, when the disparity map was generated with 11.\% and 4\% of $I_2$ respectively.}
\label{fig:VariousScenes}
\end{figure}

In addition to the middlebury dataset, we tested our algorithm on various stereo pairs,
some captured by UCSD Vision and Graphics Laboratories \cite{Zwicker:2006:AAD} and some downloaded from Flicker and rectified manually \cite{Basha:2011:GCS}.
Since the ground truth disparity is unavailable for those images, we measure the PSNR between the original $I_2$ and the warped and enhanced result $\hat{I_2}$.
Figure \ref{fig:VariousScenes} shows $I_2$ in the first column for various datasets, and the enhanced $\hat{I_2}$ for two different compressions: in the 2\textsuperscript{nd} column the disparity map is calculated with only 11.1\% of the pixels of $I_2$; in the rightmost column, only 4\% of $I_2$'s pixels were transmitted to the host.

\section{Conclusions}

We proposed an algorithm for recovering depth using less than two images, in order to reduce the communication costs.
Specifically, we have shown that Joint Bilateral Filter (JBF) offers a simple and attractive way to compress correlated images that can not communicate with each other, as is the case in practical scenarios.

In our experiments, one camera sends a full image to the host to serve as a reference, while the other camera sends as little as $2.1\%$ pixels to the host. The host can then use JBF to recover an initial depth map and use it, together with the reference image to recover the sampled image.

Our algorithm is quite fast, since both the Bilateral filter's complexity and the Non Local aggregation's complexity are linear in the image size and the disparity search range. This is significantly more efficient than previously suggested distributed source coding schemes.

There is a trade off between the amount of data transmitted and the quality of the reconstruction. This paves the way to camera arrays that can adjust the number of pixels sent to the host based on the particular bandwidth of the host and still produce a depth image that, in turn, can be used to synthesize the encoded images. In scenarios where a feedback channel exists, the errors due to occlusions can be significantly minimized. The algorithm is efficient and can be made to run at several frames per second.

\bibliographystyle{IEEEtran}
\bibliography{IEEEabrv,main}

\end{document}